# Leading Across the Spectrum of Human-AI Relationships

A Conceptual Framework for Increasingly Heterogeneous Teams

Alejandro R. Jadad, MD DPhil LLD

Research Professor (Adjunct), Keck Medical School, University of Southern California &
Founder, Vivenxia Group, LLC; Los Angeles, California, USA

(aj_492@usc.edu or ajadad@gmail.com for correspondence)

## Abstract

What shapes a consequential decision when human and artificial intelligence work on it together? The answer is becoming harder to see. A decision may look human-led after AI has set the frame, or appear automated while human judgment still carries decisive force. This paper offers a leadership-facing spectrum to see those relationships within a bounded mandate: Pure Human, Centaur (human-dominant, with AI in the loop), Co-equal, Minotaur (AI-dominant, with humans in the loop), and Pure AI.

The spectrum asks where leadership work occurs: who frames the problem, who redirects the work, and who can answer for what follows. The five positions are landmarks that help leaders recognize configurations as they layer, drift, or change in a single decision. The central risk is misrecognition: leaders may keep a human-centered story in place after decision-shaping authority has shifted elsewhere. They may believe oversight remains meaningful when it has become ceremonial, or keep humans in the loop when their involvement could make the decision worse. The framework introduces co-adaptability, the capacity of a configuration to improve as human and non-human participants adjust together, and places it within heterogeneous teaming, where participants may vary by number, substrate, model architecture, capability, speed, memory, and form of participation.

The aim is practical: to help strategic leaders and those designing or deploying AI systems recognize the configuration at work, notice when it shifts, and judge whether it fits the decision before them. These configurations will shape how power, responsibility, and trust are distributed in organizational life. Whether the futures they help create remain governable and worth inhabiting will depend on leaders who can see, early enough, where and how consequential decisions are actually being shaped.

**Keywords:** Conceptual framework; strategic leadership; human-AI relationships; heterogeneous teams; hybrid intelligence; organizational decision-making

# Introduction

Strategic leaders need a way to tell what kind of human-AI relationship is shaping the decision in front of them. They need to know whether that relationship fits the task, and who or what is really steering the work. As AI systems enter more organizational decisions, alone, together, and alongside human teams, that need will grow.

Without such clarity, leaders may continue to assume that judgment remains human after decisive influence has already moved elsewhere. They may also lean on AI where human judgment should remain dominant. Accountability can appear intact because a named person still approves the decision, even when much of the work that shaped it has been distributed across systems, routines, and interfaces that no one can fully describe. Research on hybrid intelligence, human agency in AI-supported organizational decision-making, and AI's effects on top-management work points to the same gap: the vocabulary for naming these conditions remains underdeveloped (1–3).

A pragmatic rapid scoping review (Appendix) suggested that the surrounding literature is substantial and dispersed across adjacent fields. One review of AI and organizational leadership maps work on decision support, changing leadership roles, and ethics or sustainability, and concludes that the field remains fragmented (4). Another review of AI-human collaboration in decision-making describes a rapidly growing literature distributed across several conceptual streams and proposes a new organizing structure (5). Work on top managers and AI shows similar attention to how executive capabilities and strategic work are changing (6). What remains scarce is a simple way for leaders to recognize which relationship pattern fits a given decision, and to notice when that pattern shifts before the shift hardens into habit.

The framework responds to that need. It pulls together existing insights into a practical vocabulary for leading increasingly heterogeneous teams.

# The spectrum of human-AI relationships

## What this framework asks leaders to notice

Leaders working with AI today often find themselves reaching for words that do not quite fit. "Tool," "assistant," "collaborator," and "partner" each carry assumptions that can obscure the range of arrangements through which consequential decisions now pass. The terms introduced below are meant for seeing, not for labeling.

The *framework* is what this paper offers as a whole: a way for leaders to recognize and interpret how human and AI participate in consequential organizational decisions.

The *spectrum* is the framework's organizing heuristic. It arranges human-AI relationships according to how leadership functions are distributed among participants: who frames the problem, who generates options, who challenges them, who redirects or decides, and who remains answerable for the result.

A *position* is a recognizable point on the spectrum: a familiar pattern in the distribution of leadership functions. The five positions described below are landmarks. Real arrangements often sit between them, layer them, or move among them.

A *configuration* is what is actually happening in a specific case. It is the arrangement a leader enters, creates, or inherits. Positions are abstractions; configurations are what leaders encounter in specific cases.

A *bounded mandate* is the specific decision, task, process, function, or responsibility within which the framework is being applied. Most organizations contain many bounded mandates at once, each with its own configuration. The framework does not assume that an entire organization occupies a single position. It assumes that a specific decision or function can be located.

*Heterogeneous teaming* is the broader setting that makes this framework increasingly necessary. It refers to teams whose participants differ materially in substrate, cognitive architecture, capability, temporality, or mode of participation. The more heterogeneous the team becomes, the more useful it is to have a vocabulary for the relationships within it.

*Co-adaptability* is the capacity of a configuration to adjust and become stronger under changing conditions as participants learn, recalibrate, and remain effective together over time. It is treated here as a property of the arrangement itself.

## The five positions

The premise is simple: look at where the leadership work is being done. Who frames the problem. Who generates options. Who challenges them. Who redirects the work. Who decides. Who answers for the outcome. In some configurations, these functions remain concentrated in human participants. In others, they shift toward non-human participants. In many consequential settings, they are distributed in ways that leaders may not immediately recognize.

This way of seeing builds on adjacent scholarship on human agency in AI-supported organizational decision-making, hybrid intelligence systems, AI and organizational leadership, and AI-human collaboration in decision-making (2–5). The contribution here is to convert those adjacent streams into a simpler leadership-facing spectrum.

The spectrum runs from human-dominant to AI-dominant configurations. At one end, bringing AI into the operative loop would predictably worsen the quality, safety, legitimacy, or appropriateness of the decision. At the other end, keeping humans in the operative loop would predictably worsen performance or outcomes. Between those poles, human and non-human participants share the work in different patterns of dominance, dependence, complementarity, challenge, and adjustment. Where a configuration sits matters because it changes what judgment requires, where authority lives, how performance is understood, and who can meaningfully answer for what follows.

### *Pure Human*

In these configurations, AI remains outside the operative loop because its inclusion would make the outcome worse. Human judgment is part of what the decision must preserve.

Some decisions depend on visible human responsibility. Others require moral, relational, or contextual judgment that would be damaged by algorithmic participation. AI may add false precision, flatten morally relevant nuance, weaken the meaning of the exchange, or displace the very human presence the situation requires. Keeping AI out is a design choice made to protect what matters.

Examples include a board chair explaining a grave decision to families after a catastrophe, a senior clinician discussing irreversible options with a cancer patient and loved ones, or a chief executive directly owning the decision to close a plant and address affected workers.

### *Centaur*

Here, humans lead while AI works inside the loop. The human participant frames the problem, directs the work, and retains final authority. AI contributes analysis, pattern recognition, simulation, option expansion, or structured challenge, but the work remains human-led.

The centaur metaphor has been used formally to describe productive human-algorithm combinations in which human judgment and machine capability are brought together under human direction (7). In organizational settings, the Centaur position describes cases where AI improves performance without taking over the frame.

Examples include a chief executive using AI to generate strategic scenarios before choosing a direction, a hospital leader using AI-supported forecasts to allocate capacity while retaining final judgment, or an investment committee chair using AI to stress-test options while keeping the authority to decide.

*Co-equal*

These configurations are those in which both human and non-human participants are genuinely shaping the work, not merely contributing to it. *With-ness* is the term offered here for what makes these configurations distinctive: the irreducible working-together that the existing language of collaboration, partnership, support, and synergy keeps missing. The term does not mean equal legal authority, final accountability, moral standing, or personhood. In most organizations, formal responsibility will remain with the humans involved. What makes a configuration Co-equal is something more specific: each participant can change where the work is going without being reduced to a helper, an input, or a rubber stamp.

Adjacent work on hybrid intelligence and human agency in AI-supported organizational decision-making (2,3) frames the surrounding terrain; the claim here is narrower. Co-equal names configurations in which each participant can materially redirect the work: change the frame, alter what counts as evidence, affect the next step, and be registered by the other participant as more than a source of input. These tests separate Co-equal from arrangements that can look similar from the outside, especially a strong Centaur in which AI contribution is powerful but the human still holds the frame and final direction. The AI may be influential without being able to redirect. Co-equal requires redirection in both directions.

A concrete example: a human researcher and an advanced non-human reasoning system working together over weeks to develop lines of inquiry across complex data. The researcher brings judgment about which questions matter, which framings will hold up, and which directions are worth the institution's time. The non-human system brings reach and pattern recognition that the human cannot match alone. As the work goes on, the human revises what counts as a promising line as the system surfaces things the former did not expect. The non-human system adjusts what it offers and how confidently it offers it as the human rejects or reframes its proposals. Neither could have done the work alone. Neither was helping the other. They were genuinely working together, and they were each changing as a result.

*Minotaur*

In these configurations, AI drives the operative logic while humans remain inside the loop. Human participants may supervise, calibrate, handle exceptions, preserve legitimacy, maintain a formal override, or carry accountability in institutional terms. Their presence remains important, but it is no longer dominant.

The metaphor has been formalized in work on manned-unmanned teaming, where control and operational centrality shift toward the non-human side (8). In organizations, the Minotaur position names a pattern that is common and easy to miss. Leaders may believe human oversight remains strong when humans have become constrained participants inside an AI-led configuration.

Examples include an executive committee relying on an AI system that sets the operative logic for capital allocation while leaders intervene mainly in exceptional cases; a hospital using AI triage that determines default prioritization while clinicians retain limited override; or a chief risk officer supervising an AI-driven compliance system whose outputs largely determine which cases are escalated, investigated, or acted upon.

*Pure AI*

Here, non-human systems hold the relevant leadership functions because human participation in the operative loop would make outcomes worse. Human input may slow the process, introduce inconsistency, reduce reliability, add cognitive bias, or degrade performance relative to an AI-only configuration. What defines the category is the judgment that human inclusion would weaken the outcome within the bounded mandate.

This position matters where autonomous systems are faster, cheaper, safer, more reliable, or more effective than human-led arrangements for a specific task, even when their logic remains only partly interpretable. As such systems spread, they raise questions about displacement, institutional redesign, governance at a distance, and the conditions under which human exclusion can be justified.

Examples include real-time high-frequency execution in financial markets, where human reaction times are too slow for the operative mandate; autonomous cybersecurity systems that must detect and respond faster than human teams can act; and large-scale infrastructure optimization systems where human intervention would predictably reduce speed, stability, or overall system quality.

# Leadership across moving configurations

A decision can change shape without anyone noticing. It may begin with a human framing the problem, deepen through genuine human-AI exchange, and end inside an AI-led process with limited human override. From a distance, the decision still looks human-led. Inside the work, authority has moved. The practical issue is which position the work resembles now, and what would show it shifting.

The five positions are landmarks. Many configurations sit between them. Several can appear inside the same decision: a board may keep the first judgment in human hands, explore options through Co-equal exchange, and hand execution to a system few people can challenge directly. The same spectrum can orient a dyad, a committee using one model, one person with several systems, or swarms of participants. Leaders need to see the dominant pattern when the wider decisional architecture is too tangled to hold in mind at once.

The middle of the spectrum carries the greatest risk of misrecognition. Co-equal work is delicate because each participant must be able to change the course of the work while preserving the exchange that makes that change valuable. Centaur asks whether the human still directs. Minotaur asks whether human presence still has force or has become ceremony. Co-equal is where leaders most often lose track of what they are sustaining. Pure Human and Pure AI make the boundary more visible, but visibility does not remove the need for justification and vigilance.

Influence and answerability do not always travel together (2,3). That gap is why movement usually comes before any formal change. An organization can drift from Centaur toward Minotaur as the system gets more capable and human challenge thins out. Co-equal work tilts when one side starts setting the frame or deciding what counts as evidence. Meanwhile titles, committees, and sign-offs stay stubbornly human-centered.

Some movement improves the work. A leader in a Centaur configuration may repeatedly override an AI system in a particular class of cases. If the system adjusts its thresholds and outputs in response, and the leader refines the heuristics used to judge those outputs, the configuration becomes co-adaptive. The change is reciprocal. You can see it in the work itself. In Co-equal settings this kind of mutual tuning matters more. Without continuing adjustment, with-ness becomes harder to sustain, and the configuration tends to slide toward dominance by one side or hollow out into ritual.

When the team itself loses its familiar shape, recognition becomes harder. Teams differ by expertise and style; they now also differ by substrate, temporality, and memory. Leaders may work with specialized systems, AI swarms, and layered setups in which different participants shape different phases of the same decision. As complexity grows, the question becomes unavoidable: who actually sets direction and answers for it?

Organizations carry pressures that make some movements easier than others. In profit-driven settings, speed and scale reward handing more initiative to non-human systems. Institutions that answer to voters or public legitimacy tend to keep human agency visible, even when it slows things down. Where authority rests on professional expertise and trust, AI-dominant arrangements meet resistance. In high-liability work, teams sometimes hide an AI-led core beneath a human sign-off sheet. These patterns recur across research on AI and leadership (1,4–6). They create regions of the spectrum that are deceptively easy to enter but hard to name clearly once inside.

Strategic leaders will face arrangements they design and arrangements they inherit. Many of them will be noticed only after movement has begun. Some will exceed current categories. The framework remains useful then because the question survives when the map does not.

The adoption question is too small. The work is messier: recognizing the configuration already in operation, testing whether it fits, watching for drift, and deciding, often without clear evidence, whether reciprocal adjustment is holding the work together or pulling it apart.

## Leadership-level lines of inquiry

The framework is meant first as a way to interpret what is happening. It also opens questions that can be studied.

Do leaders who can name the configuration they are working in make clearer, more accountable decisions than those who adopt AI without that clarity? Differences might appear in how accountability is allocated, how well confidence is calibrated, whether trust holds inside the team, and how the team adapts when conditions change.

Do different parts of the spectrum produce different patterns of trust, legitimacy, innovation, and error handling? The point is to map trade-offs.

Does visible mutual adjustment over time predict resilience and learning better than the usual proxies for AI readiness, such as infrastructure, training, or data governance? The test is whether the quality of mutual tuning matters more than the tools themselves.

Does pressure shape drift? Firms pushed hard on costs may slide toward greater non-human initiative without deciding to. Institutions facing political or reputational scrutiny may work harder to keep human agency visible. The issue is whether the move was chosen or simply happened.

Does the ability to shift configurations deliberately help under uncertainty? Leaders who can move with context, stakes, and aims may do better than those who stick to one default. The question is whether configurational agility itself becomes an advantage when human-AI participation keeps changing.

The possibility of a null result should remain open. Once stakes, task complexity, and constraints are accounted for, there may be no consistent performance difference across configurations. Stating that possibility keeps the framework honest.

# Discussion

The framework offered here gives strategic leaders a way to recognize human-AI relationships at the level where consequential decisions are framed, directed, and made. Its contribution is intentionally modest. It gathers insights from adjacent work on hybrid intelligence, distributed agency, AI-supported decision-making, and organizational leadership, then turns them into a practical vocabulary for increasingly heterogeneous teams.

The five positions mark regions of a continuous spectrum. A real configuration may sit between positions, combine them within one decision, or move across them as the work develops. The value of the spectrum lies in making those movements visible before older language hides them under familiar names.

The framework remains pragmatic. It does not depend on settled claims about AI consciousness, moral standing, or legal personhood. It asks a narrower question: how are leadership functions being distributed within the bounded mandate? That question can be useful even when recognition is partial, retrospective, or contested. In many settings, seeing the pattern may matter before anyone has the authority or capacity to change it.

The bounded mandate is also the framework's limit. It is designed for specific decisions, tasks, processes, functions, and responsibilities. It is not a theory of whole organizations, societies, or human-AI futures. That limitation is what allows it to travel. The same question can be asked inside a board decision, a clinical workflow, a capital-allocation process, a regulatory system, a platform, a network, or a swarm.

The hardest questions will arise as configurations become more adaptive and more heterogeneous. Co-adaptability, multi-agent coordination, and mixed human and non-human participation will not always arrive in recognizable forms. Some arrangements will be designed, others inherited, and still others discovered only after authority has already shifted. The diagnostic table gives that question a practical form: name the configuration in operation, see where misrecognition is most likely, and decide what the moment requires.

The broader implication is already visible. Leadership will depend increasingly on judging where human and non-human participants belong within particular mandates, what roles they should play, and when those roles have begun to shift. Misrecognition has consequences: misplaced confidence, symbolic oversight, blurred answerability, unnecessary human exclusion, or human presence preserved after it has lost force. The framework is meant to help leaders see those patterns early enough to judge them.

Human-AI relationships are entering the core of how consequential decisions are shaped, justified, and lived with. Their design will influence power, responsibility, and trust. Leaders who see this clearly will help determine which futures remain governable and worth inhabiting.

**Figure.** The spectrum of human-AI relationships

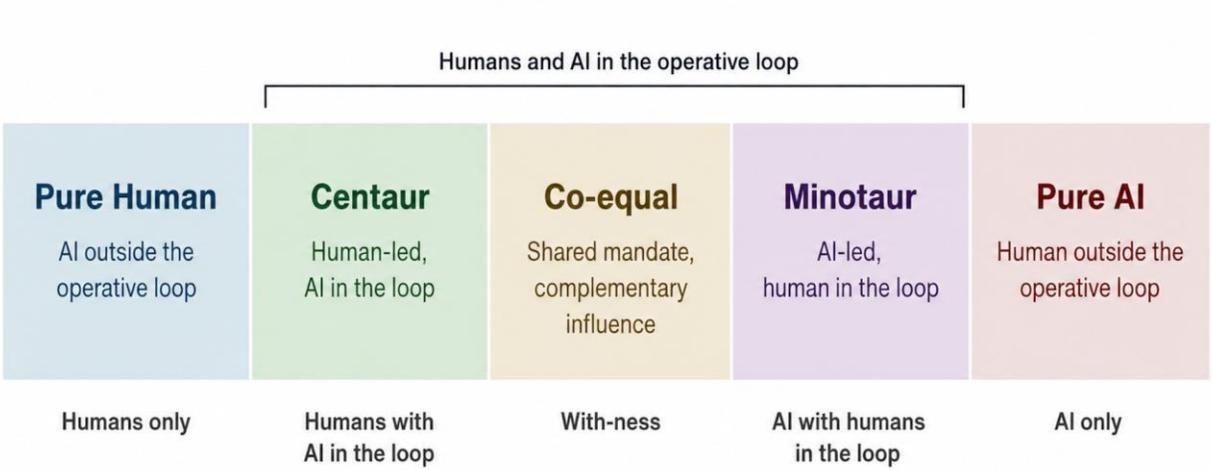

*The five positions are illustrative waypoints along a continuous spectrum. Real cases may occupy intermediate, layered, or mixed configurations.*

**Table.** Practical diagnostic guide to the spectrum of human-AI relationships

| Position | How to recognize it | Main risk if misread | Leadership task | Pressure toward it |
|---|---|---|---|---|
| **Pure Human** | AI is kept outside the operative loop because its inclusion would worsen the outcome. Human presence is itself part of what makes the decision acceptable or effective. | Treating a protected human domain as if it were open to optimization, efficiency gains, or automation. | Justify clearly why human-only judgment is preferable here and protect that boundary deliberately. | Moral, relational, professional, or legitimacy demands. |
| **Centaur** | Humans lead while AI is actively in the loop. AI improves analysis, options, or challenge, but final framing and authority remain clearly human. | Mistaking AI support for neutral assistance and drifting into hidden dependence. | Preserve meaningful human direction while using AI to improve performance. | Performance pressure without willingness to cede authority. |
| **Co-equal** | Human and non-human participants can each redirect the work, reframe the problem, shift what counts as evidence, and affect the next step. | Treating with-ness as mere support, or missing the drift toward hidden human or AI dominance. | Sustain the conditions that allow mutual redirection without collapse into dominance or ritual. | High-complexity work requiring different strengths to operate together. |
| **Minotaur** | AI drives the operative logic while humans remain in the loop for oversight, exceptions, legitimacy, or residual override. | Believing that human oversight remains meaningful when it has become largely symbolic. | Test whether human participation still has real agency, challenge capacity, and accountability value. | Speed, scale, standardization, and growing trust in AI outputs. |
| **Pure AI** | Humans are kept outside the operative loop because their inclusion would worsen the outcome. Human roles shift to design, governance, audit, or ex post review. | Preserving human participation for symbolic comfort when it degrades outcomes, or ignoring governance problems because performance is high. | Justify the exclusion and govern from outside the loop. | Strong performance advantages for autonomous systems. |

*The table presents dominant patterns. Real cases may be intermediate, layered, or mixed, and more than one position may be present within the same decisional context.*

**Appendix**. Pragmatic rapid scoping review: methods and yield

This paper was informed by a pragmatic rapid scoping review designed to identify recent scholarly and grey literature relevant to strategic organizational leadership and artificial intelligence. The review was conducted to support a conceptual framework paper, not to serve as a full systematic review. Its purpose was to determine whether a comparable leadership-facing relational framework already existed, identify adjacent streams of work, and situate the present framework within a fragmented field.

The review focused on literature addressing strategic organizational leadership and AI, human-AI collaboration in decision-making, hybrid intelligence systems, distributed agency, organizational decision-making, governance, oversight, and related leadership implications. Eligibility was concept-led. Sources were retained when they contributed directly to how human and non-human participation is configured, interpreted, or governed in organizational settings.

Scholarly searching included Ovid MEDLINE, Scopus, and Web of Science. MEDLINE was not retained as a central source because the yield was too clinically concentrated for the purposes of this paper. Subsequent searching focused on Scopus and Web of Science, supported by backward reference screening and forward citation scanning. Grey-literature searching covered high-yield institutional and organizational sources relevant to AI, leadership, governance, and institutional oversight, including NIST, OECD, World Economic Forum, McKinsey & Company, Deloitte Insights, PwC, BCG, MIT Sloan Management Review, and selected additional sources when directly pertinent.

Searches documented in the review record spanned March 30–31, 2026. The scholarly strategy was refined iteratively, beginning with broad combinations of leadership-centered and AI-centered terms and narrowing as results became dominated by clinically focused, governance-heavy, or otherwise adjacent literature. One documented Scopus strategy combined title terms such as "strategic leadership," "organizational leadership," "AI governance," "corporate governance," and "human control" with title/abstract/keyword terms including "artificial intelligence," "generative AI," "large language model," "AI agent," and "agentic AI." Later refinements tightened the search around the paper's core concern.

The documented yields illustrate the search process and the narrowing required. Early Scopus searches produced very large yields, ranging from 198,302 to 36,582, 15,785, 13,144, 2,810, and 1,306 records as search terms were refined. A documented Scopus query yielded 471 records. The latest refined Scopus search discussed in the review record yielded 80 citations, of which 64 were from 2022 onward and 2 were described as eligible reports. A Web of Science core search yielded 21 potentially eligible citations, of which 2 were described as potentially useful references. Exact numbers of all screened, excluded, and retained records were not fully documented and are therefore not reported here.

The review identified a sparse core literature surrounded by larger adjacent literatures. Relevant work was dispersed across AI-supported decision-making, hybrid intelligence, distributed agency, organizational governance and oversight, leadership adaptation, and organizational transformation. Broad searches repeatedly retrieved governance-heavy material, requiring iterative refinement. The retained literature supported the paper's central premise: concepts relevant to human-AI decision configurations exist across adjacent domains, but they are not yet organized into a simple leadership-facing vocabulary for recognizing how initiative, direction, oversight, and answerability are distributed within bounded mandates.

The review informed the paper by identifying the most relevant adjacent streams, confirming the absence of an equivalent leadership-facing relational spectrum in the retrieved literature, and anchoring the framework in recent work on AI and organizational leadership, hybrid intelligence, AI-human collaboration in decision-making, human agency in AI-supported organizational configurations, and the centaur and minotaur metaphors.

This review was pragmatic, rapid, and concept-led. It was not designed to exhaust all literature on AI and organizations, and it does not support PRISMA-style claims about complete retrieval or screening. Relevant studies may have been missed despite structured searching, citation screening, and targeted grey-literature review. The contribution of the paper is therefore conceptual rather than evidence-synthetic.